# Similarity-based Distance for Categorical Clustering using Space Structure

Utkarsh Nath[1], Shikha Asrani[2], Rahul Katarya[3]

*Department of Information Technology, Delhi Technological University*

*Delhi, India*

[1]`nath.utkarsh1@gmail.com`

[2]`shikha.asrani10@gmail.com`

[3]`rahulkatarya@dtu.ac.in`

*Abstract*—Clustering is spotting pattern in a group of objects and resultantly grouping the similar objects together. Objects have attributes which are not always numerical, sometimes attributes have domain or categories to which they could belong to. Such data is called categorical data. To group categorical data many clustering algorithms are used, among which k-modes algorithm has so far given the most significant results. Nevertheless, there's still a lot which could be improved. Algorithms like k-means, fuzzy-c-means or hierarchical have given far better accuracies with numerical data. In this paper, we have proposed a novel distance metric, similarity-based distance (SBD) to find the distance between objects of categorical data. Experiments have shown that our proposed distance (SBD), when used with the SBC (space structure based clustering) type algorithm significantly outperforms the existing algorithms like k-modes or other SBC type algorithms when used on categorical datasets.

## I. Introduction

Given a huge list of objects, "grouping similar objects together and separating dissimilar objects" is a fundamental problem in science and nature. In biology, we need to find similar proteins. In genetics, we want to find similar DNA sequences.This problem is called clustering. In computer science, it arises in several different contexts like image analysis, text analysis, speech processing and much more. K-means, fuzzy-c-means and hierarchical clustering algorithms are among the traditional algorithms developed for clustering [1]-[3]. Clustering algorithms like these work on numerical data. It is possible to work on numerical data because they can be visualized in space structures like Euclidean space structure. Space structure is a representation where the data can be visualized in a dimensional space. This makes data easy to work on. In k-means algorithm, the objects are plotted on Euclidean space and then distributed into clusters based on the distance of objects from the cluster centers. For categorical data, this calculation is not possible directly. For an object that has colour red and another object that has colour blue, the distance cannot be directly calculated between the two objects. Here the categorical objects can neither be plotted in Euclidean space nor the difference calculation makes any sense.

In the recent past, there has been need of techniques to cluster categorical data [4]-[7]. The technique of simply mapping categorical terms to numerical values to cluster them with traditional clustering techniques like k-means does not work. For clustering of categorical data, the representative techniques work on 0-1 distance [8],[9]. According to this, distance is 0 if the values are same, otherwise, the distance is 1. Though unoptimized, this form of distance is used in several representative popular algorithms. One of the most popular algorithms amongst them is k-modes [10].It is an improvement on the k-means algorithm, overcoming its lack of ability to cluster categorical data. K-modes follows the same pattern as that of k-means algorithm apart from the manner of analyzing data. In k-modes the cluster center is the category that has maximum frequency amongst all the categories of a feature for all objects belonging to the cluster. Several variations of k-modes have also come forth in the recent past [11]-[15].

Though k-modes and k-modes type algorithms seem to solve the problem of fitting categorical data into a traditional clustering algorithm by making minor changes, there are still points that need to be speculated. To mention a few:

- The performance of the cluster center calculation of k-modes(categorical data) in comparison to the one in k-means(numerical data).
- The validity of 0-1 distance, as it restricts us to only two possibilities of distance being either 0 or 1.Whereas the dissimilarity between objects might not actually be so.

A significant improvement would be, implementing the structure. It would help bringing the categorical data algorithms at par with numerical data algorithms in terms of performance. There have been trials with space structure in SBC [16]. Providing space structure to categorical data aids organizing it into clusters. It provides it the representation that maps it from categorical data to the Euclidean space. We have tried to improve the structure by proposing a novel distance calculation technique (SBD) to calculate the distance between two objects. Euclidean or other distances do not clearly work on the concept we have tried to incorporate with our distance.

In this paper, we have conducted experiments to compare k-modes and SBC using Euclidean, Cosine and Manhattan distances against space structure with SBD. Also, the reasoning that some data sets perform better on some set of clustering algorithms has been exploited. We have used k-means, fuzzy-c-means and hierarchical clustering to test all possibilities. We have got positive results for the SBD distance.

The following paper has thoroughly explained the proposed Space Structure for Categorical Data in Section II. Section III has the experimentation results, analysis and reasoning. It is followed by the Conclusion in Section IV with various remarks about the present and future work.

# II. Space Structure for Categorical Data

*A. Creating space structure*

We have many popular clustering algorithms like k-means, fuzzy-c-means, hierarchical, but the problem with these clustering algorithms is that they can be directly used only on numerical data. Numerical data is the type of data that can be represented on euclidian space. Categorical datasets have to converted to a form that can be represented on Euclidean space to be able to use the above mentioned clustering algorithms.
Let's consider a database of categorical data $T = (O, A)$, where $O = \{ o_1, o_2, o_3, \ldots, o_N \}$ is the set of objects and $A = \{ a_1, a_2, a_3, \ldots, a_M \}$ is the set of attributes of each object. N, M are the number of objects and number of attributes of an object respectively. $V(i)$ represents an array, the range of values that $A_{[i]}$ could take. Above mentioned notations are followed throughout the paper. In this paper, to represent the categorical data in euclidian space, we have created a space structure. Creating space structure from categorical data has two steps:

- Creating similarity matrix
- Creating distance matrix

  1) Creating similarity matrix
  First step is creation of a similarity matrix. A similarity matrix is a N x N matrix where each item in the matrix represents the similarity between the two objects. We have calculated the similarity between any two objects by using a similarity function $\phi$. We check the number of attributes that have same values for both the objects. For each object $o_i$, its similarity $S(i)$ is a vector of size N and $S[i][j]$ is the value of similarity between $i^{th}$ and $j^{th}$ object.

$$S[i][j] = \sum_{i=1}^{m} \omega(i, m, j) \quad (1)$$

where:

$$\phi(i, m, j) := \begin{cases} 1, & O(i, m) = O(j, m) \\ 0, & O(i, m) \neq O(j, m) \end{cases}$$

  2) Creating distance matrix
  After creating the similarity matrix, the next step is to create a distance matrix. The distance matrix is also a N x N matrix where each item in the matrix represents the distance between any two objects. We have calculated the distance between any two objects using a distance function ($\delta$). The distance function takes two vectors as an argument. The different distance functions ($\delta$) used are discussed further in the paper. For each object $o_i$, its distance $D(i)$ is a vector of size N and $D[i][j]$ is the distance between $i^{th}$ and $j^{th}$ object.

$$D[i][j] = \delta(S[i], S[j]) \quad (2)$$

TABLE I
DATA

| Object | Pressure | Volume | Temperature |
|---|---|---|---|
| Object1 | High | Medium | High |
| Object2 | Low | Low | High |
| Object3 | High | High | Low |
| Object4 | Medium | Low | High |

The above process can be better understood with an example. As given in Table I, let's consider a dataset of four objects and three attributes pressure, volume and temperature. Here N=4, M=3 and V(i) for each attribute is {low, medium, high}, where i varies from 1 to 3. Table II is the similarity matrix constructed from the data given above. $S[2][4] = 2$ as there are two attributes common between object2 and object4 and distance matrix given in Table III is then created from similarity matrix.

TABLE II
SIMILARITY MATRIX

| | | | | |
|---|---|---|---|---|
| **A** | 1 | 3 | 2 | 1 |
| **B** | 3 | 1 | 3 | 2 |
| **C** | 0 | 1 | 3 | 0 |
| **D** | 2 | 1 | 0 | 3 |

TABLE III
DISTANCE MATRIX

| $\delta(S[1], S[1])$ | $\delta(S[1], S[2])$ | $\delta(S[1], S[3])$ | $\delta(S[1], S[4])$ |
|---|---|---|---|
| $\delta(S[2], S[1])$ | $\delta(S[2], S[2])$ | $\delta(S[2], S[3])$ | $\delta(S[2], S[4])$ |
| $\delta(S[3], S[1])$ | $\delta(S[3], S[2])$ | $\delta(S[3], S[3])$ | $\delta(S[3], S[4])$ |
| $\delta(S[4], S[1])$ | $\delta(S[4], S[2])$ | $\delta(S[4], S[3])$ | $\delta(S[4], S[4])$ |

*B. Distance Function*

In this paper, we have conducted experiments with four types of distance functions. Three of them are well known distance functions and the fourth (SBD) is the one which we have proposed in this paper. The distance function is depicted by $\delta$.

- Euclidean distance

$$\delta(x,y) = \sqrt{\sum_{i=1}^{N}(x_i - y_i)^2} \qquad (3)$$

- Cosine distance

$$\delta(x,y) = \sqrt{2(1 - cos(x,y))} \qquad (4)$$

where

$$cos(x,y) = \frac{x^T y}{\|x\|.\|y\|}$$

- Manhattan distance

$$\delta(x,y) = \sqrt{(x_1 - y_1) + (x_2 - y_2) + \cdots + (x_i - y_i) + \cdots + (x_n - y_n)}$$

$$\delta(x,y) = \sqrt{\sum_{i=1}^{N}(x_i - y_i)} \qquad (5)$$

- Similarity based Distance (SBD)

After creating similarity matrix our next step is to create the distance matrix. The similarity matrix is N x N matrix, to find distance between two objects i and j we consider the similarity vector S[i] and S[j] as location Let's consider objects as given in table II. To find the distance between any two objects we'll first calculate the difference between the attributes. In table II we can see that both the difference between $o_1$, $o_2$ and $o_2$, $o_3$ is 1(considering only the first attribute) whereas the first being 3-2 and latter 2-1. The whole point of SBD is that objects having more in common should be closer i.e $o_1$, $o_2$(3,2) should be closer than $o_2$, $o_3$(2,1). Euclidean or Manhattan distance if used in the above mentioned condition would have assigned same value to the distance between $o_1$, $o_2$ and $o_2$, $o_3$. The magnitude of each dimension is contributed by one of the objects (one to one relationship).In case either one of S[x][i] or S[y][i] is 0 denominator becomes 1 to prevent divide by 0 error.

$$\delta(x,y) = \sqrt{\sum_{i=1}^{N} \frac{\|S[x][i] - S[y][i]\|^2}{\|S[x][i]\|\|S[y][i]\|}} \qquad (6)$$

## C. Clustering Algorithm for Categorical Data

By generating the distance matrix, we have converted the categorical data to a form on which clustering algorithms like k-means, fuzzy-c-means or hierarchical can be applied. We consider the distance matrix as N objects with each of N dimensions. In Algorithm 1 the steps of the algorithm discussed above are summarised.Code for this can be downloaded from here[23].The value K is same as the number of classes the data is divided into.

---

**Algorithm 1:** Clustering Algorithm for Categorical Data

**Input**: categorical data T = (O, A). Where O = { $o_1$, $o_2$ ,…, $o_N$ } is the set of objects and A = { $a_1$, $a_2$ ,….., $a_M$ } is the set of attributes of the categorical data.
**Output**: K Clusters
(1) Create a Similarity matrix S from categorical data T = (O, A). $\phi$ is a similarity function that return a boolean value, depending upon whether the attributes are same for the two objects or not.

$$S[i][j] = \sum_{i=1}^{m} \phi(i,m,j) \qquad (7)$$

where:

$$\phi(i,m,j) := \begin{cases} 1, & O(i,m) = O(j,m) \\ 0, & O(i,m) \neq O(j,m) \end{cases}$$

(2) Construct a Distance matrix (D) from Similarity matrix (S) using SBD as the distance function.

$$D[i][j] = \delta(S[i], S[j]) \qquad (8)$$

where:

$$\delta(x,y) = \sqrt{\sum_{i=1}^{N} \frac{\|S[x][i] - S[y][i]\|^2}{\|S[x][i]\|\|S[y][i]\|}}$$

(3) Assuming Distance matrix (D) as objects of N dimensional data, apply any of the three clustering algorithm - k-means, fuzzy-c-means, hierarchical.

---

## III. Experiment

The performance of clustering algorithm is judged by its ability to distribute the data into clusters that reflect their inherent patterns. We carry out an experimental analysis to verify the distribution given by the clustering algorithm against the required cluster distribution. The data is first converted into similarity matrix and then we calculate distance matrix from this similarity matrix. The distance metric used here is what brings the difference in performance. We have used Euclidean, Cosine, Manhattan and SBD distance. The objects are clustered according to the similarity among objects, where this matrix serves as the feature set for the objects. We have used 3 clustering algorithms k-means, fuzzy-c-means and hierarchical. We have compared each of the 4 mentioned distance matrices for each clustering algorithm and k-modes algorithm separately.

The clustering algorithms have been run 100 times and the average of these runs is calculated to get the final result.

Following are the values of Accuracy measure [17]. We have taken into account 12 datasets from the UCI repository to test the performance [18]. The basic statistics of datasets are reflected in Table IV. Missing data in the datasets have been handled by considering it as an another category of the attribute. The details of the results are given in Table V-VIII.

The best result for each dataset is in bold font and underlined.

TABLE IV
DESCRIPTION OF DATA SETS FROM UCI

| Dataset | Instances | Classes | Features |
|---|---|---|---|
| Fitting Lenses | 24 | 3 | 4 |
| Breast Cancer | 699 | 2 | 9 |
| Balloon | 20 | 2 | 4 |
| SoyBeen Small | 47 | 4 | 35 |
| HayesRoth | 132 | 3 | 4 |
| Promoters | 106 | 2 | 57 |
| Monks | 432 | 2 | 6 |
| Voters | 435 | 2 | 16 |
| Shuttle Landing Control | 15 | 2 | 6 |
| Balance Scale | 625 | 3 | 4 |
| Soybean Large | 307 | 19 | 35 |
| Fisher Order | 150 | 3 | 4 |

TABLE V
VALUES OF ACCURACY FOR K-MODES

| Dataset | Accuracy |
|---|---|
| Fitting Lenses | **<u>0.666</u>** |
| Breast Cancer | 0.911 |
| Balloon | 0.875 |
| SoyBeen Small | **1** |
| HayesRoth | 0.51 |
| Promoters | 0.650 |
| Monks | 0.590 |
| Voters | 0.86 |
| Shuttle Landing Control | 0.6 |
| Balance Scale | 0.488 |
| Soybean Large | 0.644 |
| Fisher Order | **<u>0.361</u>** |

TABLE VI
VALUES OF ACCURACY FOR K-MEANS CLUSTERING

| Dataset | Accuracy | | | |
|---|---|---|---|---|
| | SBD | Euclidean | Cosine | Manhattan |
| Fitting Lenses | 0.625 | 0.625 | 0.625 | 0.541 |
| Breast Cancer | 0.869 | 0.931 | 0.838 | 0.844 |
| Balloon | **<u>0.937</u>** | 0.8125 | 0.625 | 0.562 |
| SoyBeen Small | **1** | **1** | **1** | **1** |
| HayesRoth | 0.530 | 0.515 | 0.439 | 0.462 |
| Promoters | 0.915 | 0.886 | 0.858 | 0.783 |
| Monks | 0.513 | 0.578 | 0.578 | 0.513 |
| Voters | 0.875 | 0.873 | 0.873 | 0.873 |
| Shuttle Landing Control | 0.533 | 0.533 | **<u>0.666</u>** | 0.533 |
| Balance Scale | 0.534 | 0.6 | 0.6 | 0.464 |
| Soybean Large | 0.62 | 0.662 | 0.624 | 0.644 |
| Fisher Order | **<u>0.361</u>** | **<u>0.361</u>** | **<u>0.361</u>** | **<u>0.361</u>** |

TABLE VII
VALUES OF ACCURACY FOR F-C-MEAN CLUSTERING

| Dataset | Accuracy | | | |
|---|---|---|---|---|
| | SBD | Euclidean | Cosine | Manhattan |
| Fitting Lenses | 0.625 | 0.541 | 0.66 | 0.625 |
| Breast Cancer | 0.866 | 0.891 | 0.828 | 0.849 |
| Balloon | 0.812 | 0.687 | 0.812 | 0.562 |
| SoyBeen Small | 1 | 1 | 1 | 1 |
| HayesRoth | 0.545 | 0.477 | 0.507 | 0.477 |
| Promoters | 0.915 | 0.886 | 0.915 | 0.896 |
| Monks | **0.747** | 0.578 | 0.578 | 0.513 |
| Voters | 0.875 | 0.873 | 0.873 | 0.873 |
| Shuttle Landing Control | 0.533 | 0.533 | **0.666** | 0.533 |
| Balance Scale | 0.464 | 0.384 | 0.483 | 0.484 |
| Soybean Large | 0.634 | 0.655 | 0.620 | 0.606 |
| Fisher Order | **0.361** | **0.361** | **0.361** | **0.361** |

TABLE VIII
VALUES OF ACCURACY FOR HIERARCHICAL CLUSTERING

| Dataset | Accuracy | | | |
|---|---|---|---|---|
| | SBD | Euclidean | Cosine | Manhattan |
| Fitting Lenses | 0.625 | 0.625 | 0.625 | 0.625 |
| Breast Cancer | **0.951** | 0.939 | 0.842 | 0.851 |
| Balloon | 0.812 | 0.812 | 0.687 | 0.562 |
| SoyBeen Small | 1 | 1 | 1 | 1 |
| HayesRoth | 0.553 | 0.431 | **0.575** | 0.454 |
| Promoters | 0.924 | 0.886 | **0.943** | **0.943** |
| Monks | 0.513 | 0.578 | 0.578 | 0.513 |
| Voters | **0.878** | 0.818 | 0.875 | 0.818 |
| shuttle-landing-control | **0.666** | 0.533 | **0.666** | 0.533 |
| Balance Scale | **0.67** | 0.6 | 0.531 | 0.6 |
| Soybean Large | 0.648 | **0.665** | 0.606 | 0.637 |
| Fisher Order | | **0.361** | **0.361** | **0.361** | **0.361** |

### A. Accuracy

It is the measure of the efficiency of the algorithm. It indicates the point to which the algorithm's results are matching to the pattern of clusters required. It is the fraction of total objects that are placed correctly in their respective clusters. It can be defined by the following formula:

$$AC = \sum_{i=1}^{k} \frac{\max(n_{ij} : j \leq k')}{n} \quad (9)$$

$$n_{ij} = |c_i \cap p_j|$$

$n_{i,j}$ is the number of objects common in cluster $c_i$ and cluster $p_j$, where C={$c_0$, $c_1$, ......... $c_s$} set of the clusters formed by our clustering algorithm and P={$p_0$, $p_1$, .........$p_k$} is the actual pattern of clusters required. The sum of maximum $n_{i,j}$ for 0<=j<=k for each i where 0<=i<=s, divided by the total number of objects is the accuracy. s and k are the number of clusters in C and P cluster distributions respectively.

In the results, we can see the improvement in the value of accuracy when SBD is used. Out of the ten data sets used eight have given their best result using the SBD distance matrix. Code for this algorithm can be found in [19].

## *Conclusions*

Including the effect of similarity in the distance calculation has improved the performance of the algorithm. We have managed to map categorical data into a space structure effectively. We have achieved positive results for the Similarity-based Distance (SBD) proposed on the tested datasets.

There are still several issues that remain to be solved. One of the major concerns to be looked upon is clustering of mixed data. In real life the data available is not necessarily of one type, it is often a mix of many. We have achieved progress in terms of individually clustering a type of data, but work still needs to be done to merge these algorithms to be able to cluster this mixed data.

There are several other factors that can be worked upon to improve the performance of this algorithm further. One of the future scopes of this research work would be including weight in the distance calculation, to narrow down the sensitive features and give them more weight than others [20].

Combining this algorithm with genetic algorithms of optimization could also be tried. Objective functions of algorithms are optimized usually to improve results by a fair extent. In future, efforts can also be put into combining this algorithm with other techniques like decision trees [21] or rough sets [22],[23] to improve the overall performance.

## *References*

[1] J. Yu, "General C-means clustering model," *IEEE Trans. Pattern Anal. Mach. Intell.*, vol. 27, no. 8, pp. 1197–1211, Aug. 2005.
[2] ParulAgarwal, M. AfsharAlam andRanjitBiswas, "A Hierarchical Clustering Algorithm for Categorical Attributes, " International Journal of Innovation Management and Technology, vol. 1, pp. 189-190, 2010.
[3] Malay K. Pakhira, "A Modified k-means Algorithm to avoid empty clusters, " International Journal of Recent Trends in Engineering, vol. 1, pp. 1-8, 2009.
[4] T. Xiong, S. Wang, Q. Jiang, J. Z. Huang, "A novel variable-order Markov model for clustering categorical sequences", *IEEE Trans. Knowl. Data Eng.*, vol. 26, no. 10, pp. 2339-2353, Oct. 2014.
[5] C.-C. Hsu, S.-H. Lin, "Visualized analysis of mixed numeric and categorical data via extended self-organizing map", *IEEE Trans. Neural Netw. Learn. Syst.*, vol. 23, no. 1, pp. 72-86, Jan. 2012.
[6] N. Iam-On, T. Boongoen, S. Garrett, C. Price, "A link-based approach to the cluster ensemble problem", *IEEE Trans. Pattern Anal. Mach. Intell.*, vol. 33, no. 12, pp. 2396-2409, Dec. 2011.
[7] *Fuyuan Cao, Jiye Liang, Liang Bai, Xingwang Zhao, and Chuangyin Dang, "A Framework for Clustering Categorical Time-Evolving Data", IEEE Transactions On Fuzzy Systems, VOL. 18, NO. 5, OCTOBER 2010.*
[8] M. Alamuri, B. R. Surampudi, and A. Negi, "A survey of dis- tance/ similarity measures for categorical data," in *Proc. Int. Joint Conf. Neural Netw.*, 2014, pp. 1907–1914.
[9] G. Sudipto, R. Rastogi, and K. Shim, "Rock: A robust clustering algorithm for categorical attributes", in Data Engineering, 1999. Proceedings., 15th International Conference on. IEEE, 1999, pp. 512-521.
[10] Z. Huang, "A fast clustering algorithm to cluster very large categorical data sets in data mining," in *Proc. SIGMOG Workshop Res. Issues Data Mining Knowl. Discovery*, 1997, pp. 1–8.
[11] L. Bai, J. Liang, C. Dang, F. Cao, "The impact of cluster representatives on the convergence of the K-modes type clustering", *IEEE Trans. Pattern Anal. Mach. Intell.*, vol. 35, no. 6, pp. 1509-1522, Jun. 2013.
[12] M. K. Ng, M. J. Li, J. Z. Huang, and Z. He, "On the impact of dissimilarity measure in *k*-modes clustering algorithm," *IEEE Trans. Pattern Anal. Mach. Intell.*, vol. 29, no. 3, pp. 503–507, Mar. 2007.
[13] M. K. Ng and L. Jing, "A new fuzzy k-modes clustering algorithm for categorical data," *Int. J. Granular Comput., Rough Sets Intell. Syst.*, vol. 1, no. 1, pp. 105–119, 2009.
[14] Z. Huang and M. K. Ng, "A fuzzy *k*-modes algorithm for clustering categorical data," *IEEE Trans. Fuzzy Syst.*, vol. 7, no. 4, pp. 446–452, Aug. 1999.
[15] Z. He, S. Deng, and X. Xu, "Improving K-modes algorithm considering frequencies of attribute values in mode," in *Proc. Int. Conf. Comput. Intell. Security*, 2005, pp. 157–162.
[16] Y. Qian, F. Li, J. Liang, B. Liu, C. Dang, "Space structure and clustering of categorical data", *IEEE Trans. Neural Netw. Learn. Syst.*.
[17] Y. Yang, "An evaluation of statistical approaches to text categorization," *Inf. Retr.*, vol. 1, nos. 1–2, pp. 69–90, 1999.
[18] Archive.ics.uci.edu. (2017). *UCI Machine Learning Repository: Data Sets*. [online] Available at: https://archive.ics.uci.edu/ml/datasets.html [Accessed 9 Dec. 2017].
[19] (2017). Similarity based distance [Online]. Available:https://github.com/utkarshnath/Categorical-Clustering
[20] H. Jia, Y.M. Cheung, J.M. Liu, "A new distance metric for unsupervised learning of categorical data", *IEEE Transactions on Neural Networks and Learning Systems*, vol. 27, no. 5, pp. 1065-1078, 2016.
[21] Md.Z. Islam, L. Brankovic, "DETECTIVE: A decision tree based categorical value clustering and perturbation technique for preserving privacy in data mining," in 2005 3rd IEEE International Conference on Industrial Informatics, INDIN2005, art. no. 1560461, pp. 701-708.
[22] B.K. Tripathy, A. Ghosh, "SDR: An algorithm for clustering categorical data using rough set theory", *Recent Advances in Intelligent Computational Systems (RAICS)*, pp. 867-872, 2011.
[23] B.K. Tripathy, A. Ghosh, "Data Clustering Algorithms Using Rough Sets" in Handbook of Research on Computational Intelligence for Engineering, Science, and Business, pp. 297, 2012.